\documentclass[letterpaper, 10 pt, conference]{ieeeconf}  % Comment this line out if you need a4paper

\IEEEoverridecommandlockouts                              % This command is only needed if 
\usepackage{graphics} % for pdf, bitmapped graphics files
\usepackage{epsfig} % for postscript graphics files
\usepackage{times} % assumes new font selection scheme installed
\usepackage{amsmath} % assumes amsmath package installed
\usepackage{amssymb}  % assumes amsmath package installed
\usepackage{xcolor}
\usepackage{bm}
\usepackage{cite}
\usepackage[linesnumbered,ruled]{algorithm2e}
\usepackage{ulem} %to strike the words
\usepackage[colorlinks=true,linkcolor=black,citecolor=blue,urlcolor=blue,]{hyperref}
\usepackage[doipre={doi:~}]{uri}
\usepackage{float}
\usepackage{booktabs}
\usepackage{multirow}
\usepackage{mathrsfs}
\usepackage[utf8]{inputenc}
\usepackage[caption=false, font=footnotesize]{subfig}
\usepackage{graphicx}
\usepackage{pifont}

\newcounter{RNum}
\renewcommand{\theRNum}{\arabic{RNum}}
\newcommand{\Remark}{\noindent\textit{\textbf{Remark}~\refstepcounter{RNum}\textbf{\theRNum}: }}
\usepackage{soul}
\newcommand{\NoOne}[1]{\textcolor{red}{#1}}
\newcommand{\NoTwo}[1]{\textcolor{green}{#1}}
\newcommand{\NoThree}[1]{\textcolor{blue}{#1}}
 
\soulregister\NoOne7
\soulregister\NoTwo7
\soulregister\NoThree7
\soulregister\Remark7
\usepackage{flushend}

\title{\LARGE \bf
    Local Perception-Aware Transformer for Aerial Tracking
}

\author{Changhong Fu$^{1,*}$, Weiyu Peng$^{1}$, Sihang Li$^{1}$, Junjie Ye$^{1}$, and Ziang Cao$^{2}$%  <-this % stops a space
\thanks{$^{1}$Changhong Fu, Weiyu Peng, Sihang Li and Junjie Ye are with the School of Mechanical Engineering, Tongji University, 201804 Shanghai, China.
        {\tt\small changhongfu@tongji.edu.cn}}%
\thanks{$^{2}$Ziang Cao is with the School of Automotive Studies, Tongji University, 201804 Shanghai, China.}
\thanks{$^{*}$Corresponding author}
}

\begin{document}

\maketitle
\thispagestyle{empty}
\pagestyle{empty}

%%%%%%%%%%%%%%%%%%%%%%%%%%%%%%%%%%%%%%%%%%%%%%%%%%%%%%%%%%%%%%%%%%%%%%%%%%%%%%%
\begin{abstract}
Transformer-based visual object tracking has been utilized extensively. 
However, the Transformer structure is lack of enough inductive bias. 
In addition, only focusing on encoding the global feature does harm to modeling local details, which restricts the capability of tracking in aerial robots.
Specifically, with local-modeling to global-search mechanism, the proposed tracker replaces the global encoder by a novel local-recognition encoder.
In the employed encoder, a local-recognition attention and a local element correction network are carefully designed for reducing the  global redundant information interference and increasing local inductive bias.
Meanwhile, the latter can model local object details precisely under aerial view through detail-inquiry net. 
The proposed method achieves competitive accuracy and robustness in several authoritative aerial benchmarks with 316 sequences in total. The proposed tracker's practicability and efficiency have been validated by the real-world tests. The source code is available at \url{https://github.com/vision4robotics/LPAT}.
\end{abstract}
%To improve the robustness of trackers in dark scenes, DarkLighter
%visibility of low-light conditions,
%to light up the darkness for CNN-based trackers

%%%%%%%%%%%%%%%%%%%%%%%%%%%%%%%%%%%%%%%%%%%%%%%%%%%%%%%%%%%%%%%%%%%%%%%%%%%%
%%%%%%%%%%%%%%%%%%%%%%%%%%%%%%%%%%%%%%%%%%%%%%%%%%%%%%%%%%%%%%%%%%%%%%%%%%%%%%%%
\section{Introduction} \label{sec:intro}
Visual object tracking has received more and more attention in the past few decades. 
It facilitates numerous intelligent applications of robots as a fundamental task, such as obstacle avoidance~\cite{9340778}, self location~\cite{9457090}, and infrastructure inspection~\cite{9492856}. However, different from general object tracking, the object tracking for aerial robots faces more complicated and difficult challenges, \textit{e.g.}, rotation, deformation and fast motion~\cite{9445732}. Moreover, limited computation resources on aerial robots and requirements for real-time tracking call for a faster and more robust tracker.

The current trackers\cite{Li2019CVPR, chen2020siamese, bertinetto2016fully,li2018high, guo2020siamcar,zhao2020siamese} adopt ingeniously designed convolutional neural network (CNN) structure to extract local information of correlation feature map effectively. However, the CNNs' inherent shortage of modeling long-range dependencies weakens the tracker's capability to search global potential objects under aerial view. Additionally, these recent works need to stack enough layers for expanding the receptive field, which consequently causes large consumption of the computation resource.
\begin{figure}[t]
      \centering
      \includegraphics[width=0.48\textwidth]{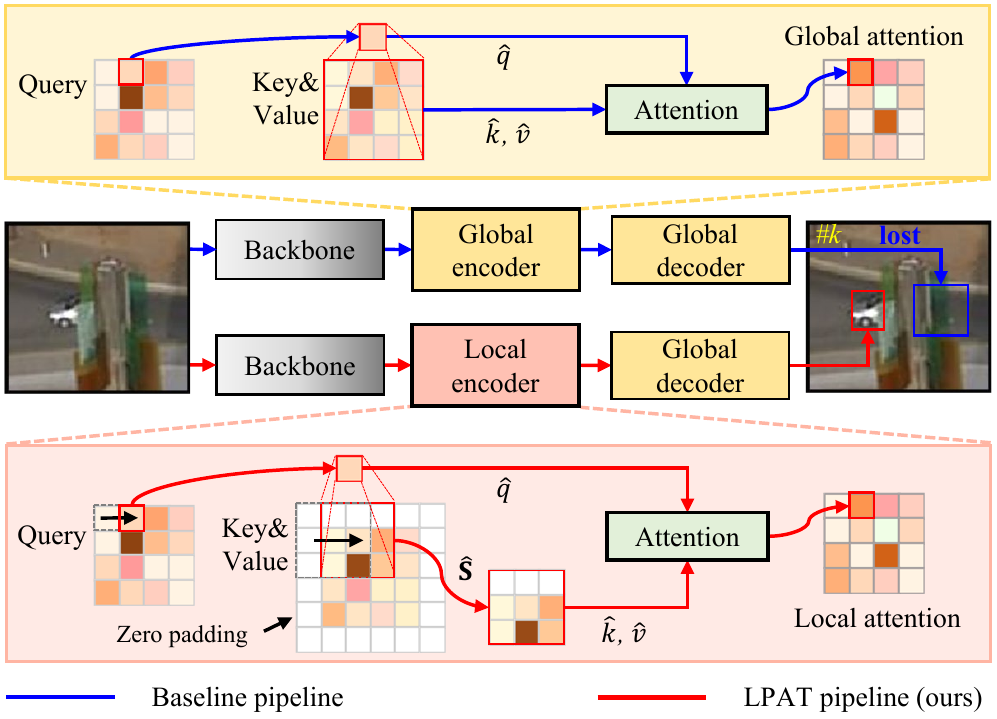}
      \setlength{\abovecaptionskip}{-12pt}
      \caption{Comparison of the baseline and LPAT. Local attention is adopted in LPAT to make the tracker focus on local details. The local-to-global architecture exploits the representational ability of Transformer further. $\hat{q}$ denotes a token of the sequence Query. $\hat{k}$ and $\hat{v}$ denote the sequences of Key and Value of the local scope $\hat{\textbf{S}}$ respectively. The proposed tracker with the local inductive bias further gains stronger robustness.}
      \label{head1}
      \vspace{-20pt}
\end{figure}

%中间是两个结构对比，两边放具体内容。caption 解释两者区别：

To address the shortages of traditional CNN-based trackers, Transformer~\cite{vaswani2017nips} is introduced into the trackers~\cite{Chen2021CVPR,yan2021learning} to increase the capability to model long-range dependencies. Benefiting from gathering global information, they can achieve comparable accuracy when adopting a backbone with fewer layers, compared with the fully CNN-based trackers~\cite{cao2021hift}. However, the typical Transformer adopted in recent works~\cite{cao2021hift,yan2021learning,Chen2021CVPR} lacks intrinsic inductive bias to deal with objects at various scales in modeling local visual structures like CNNs. Under this limitation, the Transformer encoder in a Transformer-based tracker introduces background interference and cannot effectively model the local details of the object when expanding the receptive field with global attention. To address these problems, a special construction method with more local inductive bias is needed for the Transformer-based tracker which will help capture more local details for tackling challenges, \textit{e.g.}, rotation, deformation and fast motion. 

Recent works in other fields have paid attention to the inductive bias limitation and attend to increase it in the Transformer-based structure. DeiT~\cite{touvron2021training} proposes a distilling knowledge training method from CNNs to Transformers, which requires a well-trained CNN model as the teacher to introduces the CNNs' local inductive bias but consumes extra training costs. ViTAE~\cite{xu2021vitae} combines CNNs and multi-head self-attention but lacks sufficient exploration of the overall Transformer structure's inductive bias. Limiting self-attention into a local region, ViL~\cite{zhang2021multi} adopts the overlapped windows as convolutions to process all the tokens, but lacks local finer details modeling.

Addressing the limitations above, this work constructs a novel Transformer-based aerial tracker with a local-to-global mechanism, in which two novel local perception-aware blocks are carefully designed. The two blocks are set in two parallel branches to represent local elements which further enhances the network's perception of local details. As compared with the typical Transformer-based tracker in Fig.~\ref{head1}, the proposed framework exploits the local inductive bias of the Transformer further to cope with extremely complex conditions.

%Considering to reduce background interference, the network shown in Fig. \ref{head1} adopts local-recognition Transformer to encode multi-level cross-correlation information as potential targets, decode final match results with global-detection Transformer and directly output the final set of predictions.
% The proposed local perception-aware Transformer (LPAT) siamese tracker can strengthen the acquisition and differentiation of local details, and increase the efficiency and accuracy of tracking. This work also establishes a novel information processing mechanism to modify all-over Transformer encoder-decoder which enhances the encoder's modeling ability while retaining the decoder's global-information processing capability.
The main contributions of this work are listed below:

\begin{itemize}

\item A novel Transformer-based tracker with local-modeling to global-search information processing mechanism is proposed to compensate for detailed feature modeling of the Transformer modules and retain the powerful processing capabilities for global information. 
\item A local-recognition attention block ($\rm LRA$) is designed to fuse and encode features efficiently with local receptive field. The proposed attention mechanism introduces the local inductive bias to the attention block and improves the tracker's capability to capture more local details.
\item A local element correction network ($\rm LEC$) is constructed to compensate for the detailed feature modeling of Transformer. This network can enhance the local representation by fusing multi-scale elements.

% \item A contrast loss is proposed to strengthen the discrimination of local details. Based on this loss, our tracker can learn from the sample to further distinguish the details.
\item Extensive experiments on the representative benchmarks and real-world tests show the proposed method's out-performance upon SOTA approaches fairly, which demonstrates LPAT's efficiency and robustness.

\end{itemize}

\section{Related Works}

%%%%%%%%%%%%%%%%%%%%%%%%%%%%%%%%%%%%%%%%%%%%%%%%%%%%%%%%%%%%%%%%%%%%%%%%%%%%%%%%
\subsection{Trackers for Aerial Object Tracking}

% Correlation filter-based trackers have been proved their efficiency in aerial object tracking. But correlation filters cannot maintain sufficient robustness~\cite{correlation}. 
Siamese network-based trackers~\cite{Li2019CVPR,  chen2020siamese, bertinetto2016fully, li2018high,guo2020siamcar,zhao2020siamese} adopt offline-trained CNNs to extract deep features of both template frame and searching frame and search the best matching region after feature maps correlation, which further exploits the local representation than correlation filter-based trackers\cite{li2020autotrack,huang2019learning,wang2018multi,li2018learning}. 
L. Bertinetto \textit{et al.}~\cite{bertinetto2016fully} propose the first model which incorporates feature correlation into a Siamese framework.
B. Li \textit{et al.}~\cite{li2018high} combines a Siamese network with region proposal networks for more efficient classification and more accurate bounding boxes prediction.
SiamCAR~\cite{guo2020siamcar} achieves a precise performance with a simple anchor-free framework. 
SiamRTU~\cite{zhao2020siamese} not only crafts a location module subnetwork for locating the target, but also explores the application of various model update mechanisms.
SiamAPN~\cite{fu2021onboard} exploits the anchor-free network further in Unmanned Aerial Vehicle (UAV) tracking task and gains promising performances on several typical benchmarks.
Naturally, local pixels are more likely to be correlated in images, which suggests that CNNs have an inductive bias in modeling local dependencies. 
However, trackers with fully CNN-based architecture have to adopt a deeper network and face the limitation in  efficient long-range dependencies modeling.
Such inductive bias in CNN is not sufficient but important for Transformer-based aerial tracking. This work introduces the CNN-like inductive bias to a Transformer-based tracker by cross-module local-to-global mechanism and precise local details modeling.
%%%%%%%%%%%%%%%%%%%%%%%%%%%%%%%%%%%%%%%%%%%%%%%%%%%%%%%%%%%%%%%%%%%%%%%%%%%%%%%%
\subsection{Transformer Tracking}

Transformer was first introduced by A. Vaswani \textit{et al.}~\cite{vaswani2017nips} and applied in machine translation and is recently introduced into the domain of computer vision. Vision Transformer has been proven its efficiency in various vision tasks~\cite{esser2021taming,dosovitskiy2020image,chen2021crossvit,Zheng_2021_CVPR}. Especially in the field of object tracking, Transformer-based trackers with global attention blocks demonstrate their powerful global dependencies modeling capabilities. TransT~\cite{Chen_2021_CVPR} achieves promising accuracy by replacing the standard correlation layer with cross-attention Transformer block. STARK~\cite{yan2021learning} gathers the global space and temporal information for end-to-end regression and classification, and experiments illustrate its preferable performance. To design a Transformer tracker for aerial platforms, HiFT~\cite{cao2021hift} proposes a lightweight Transformer architecture to fuse features and validate its method's efficiency in real-world tests.
TCTrack~\cite{Cao_2022_TCTrack} exploits the temporal prior knowledge further for both feature extraction and  similarity map refinement. However the current Transformer trackers all adopt the typical Transformer structures which is limited in modeling local details, and it is not enough when trackers only learn the inductive bias from data implicitly. 
% This work also designs Transformer architecture for tracking. Unlike the previous Transformers for tracking, the proposed Transformer is able to model local details more effectively and maintain its efficient processing capabilities for global information.
\begin{figure*}[t]
       \centering
       \includegraphics[width=1\textwidth]{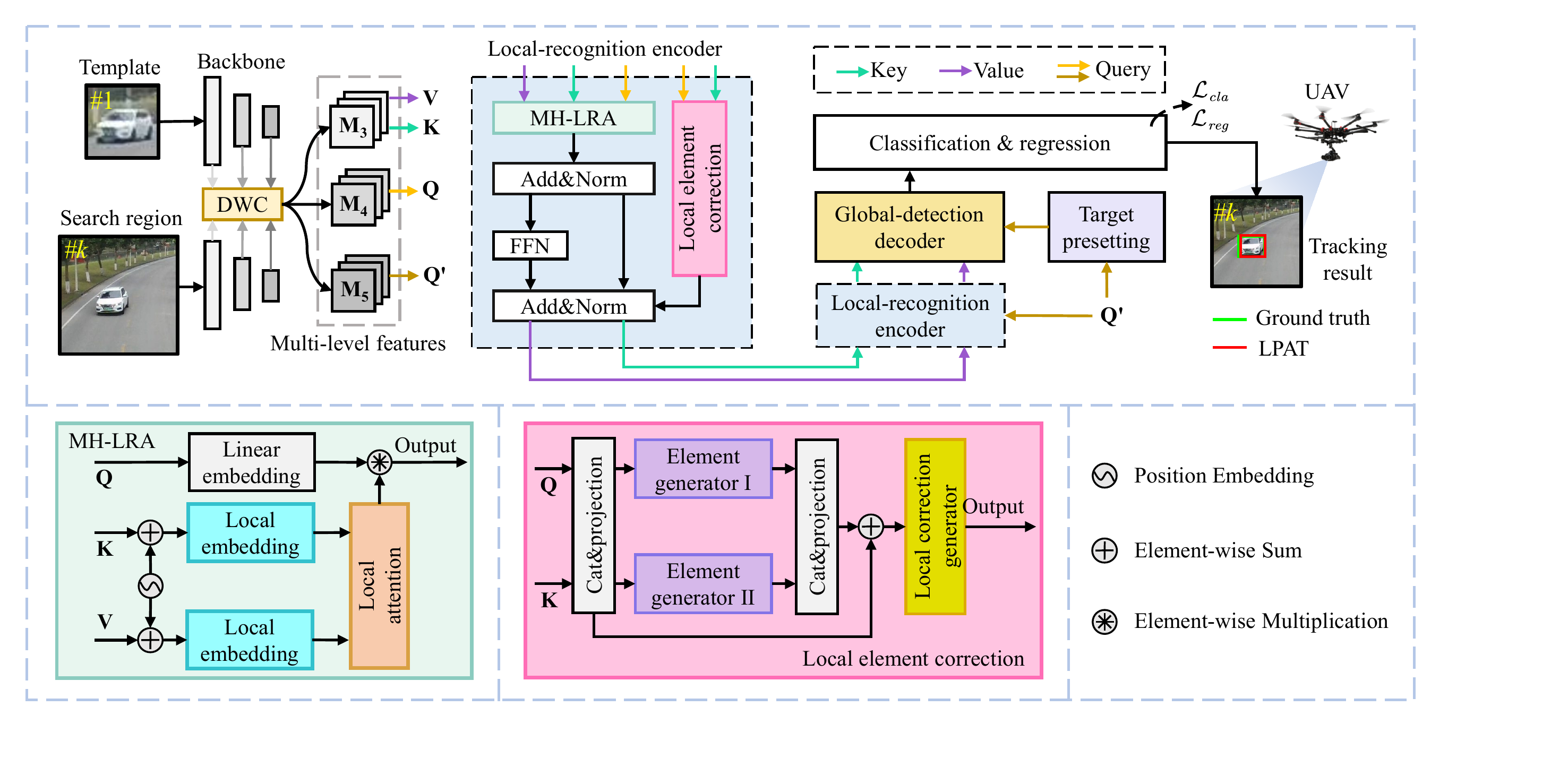}
       	\setlength{\abovecaptionskip}{-12pt}
       \caption{The proposed framework. Sub-graphs illustrate the specific structure of the modules corresponding to the colors respectively. DWC denotes the deep-wised correlation layer. MH-LRA denotes the multi-head local-recognition attention block. $\rm FFN$ is denoted as the feed-forward network. Each block is followed by a normalization layer. Template image and Searching image are input to the weight-shared backbone. The output feature maps of the last three layers operate DWC in pairs and reshape to 3 sequences of tokens, $i.e.$, $\textbf{M}_3, \textbf{M}_4, \textbf{M}_5$. $\textbf{Q}$, $\textbf{K}$, $\textbf{V}$ and $\textbf{Q'}$ represent the different input variables for whole Transformer framework. Each encoder layer is adopted to fuse features at different depths. Through Transformer decoder and CNN prediction head, the final result is selected by the regression scores. (Best viewed in color)}
       \vspace{-12pt}
       \label{zt}
\end{figure*}

%%%%%%%%%%%%%%%%%%%%%%%%%%%%%%%%%%%%%%%%%%%%%%%%%%%%%%%%%%%%%%%%%%%%%%%%%%%%%%%%
\subsection{Inductive Bias of Transformer}

Achieving promising results in vision tasks, inductive biases of the typical Transformer are exploited further. Aiming to reduce the number of datasets and speed up training, DeiT~\cite{touvron2021training} adopts knowledge distillation from CNNs to Transformers, which facilitates the training of vision Transformers by introducing the inductive bias from CNNs. On the other hand, the inductive biases required for the vision task cannot all be learned from training. Recently, some works~\cite{Liu2021swin,zhang2021multi,chu2021twins}, introduce the intrinsic inductive bias like CNNs into the attention of Transformer directly. To exploit the locality and global dependencies further, ViTAE~\cite{xu2021vitae} achieves effective representation by stacking convolutions and attention layers sequentially. Regrettably, the task-required inductive bias of Transformer is becoming an important direction in several vision tasks but not further exploited in aerial object tracking. This work firstly attempts to systematically augment the Transformer's local inductive bias to exploit cross-module local-to-global mechanism for aerial object tracking. 

%%%%%%%%%%%%%%%%%%%%%%%%%%%%%%%%%%%%%%%%%%%%%%%%%%%%%%%%%%%%%%%%%%%%%%%%%%%%%%%%

\section{Methodology}

The presented method shown in Fig. \ref{zt} is constructed with a weight-shared backbone to extract features, a Transformer module to process features, and a prediction head to predict the location of the object.

% \begin{figure}[t]
%       \centering
%       \includegraphics{graphic/3map.pdf}
%       \caption{($a$) and ($b$) represent the standard attention map and the local attention map respectively. The local receptive field is defined on 2D feature map.}
%       \label{map}
% \end{figure}

%%%%%%%%%%%%%%%%%%%%%%%%%%%%%%%%%%%%%%%%%%%%%%%%%%%%%%%%%%%%%%%%%%%%%%%%%%%%%%%%
\subsection{Local-Recognition Attention} 

In order to eliminate background interference and introduce the local inductive bias, this work proposes a novel local-recognition attention block. Different from the attention of typical Transformer modules, $\rm LRA$ adopts the local receptive field to eliminate background interference and reduce the time and space complexity to $O(n)$. Convolution layers for embedding queries and keys are deployed to explore the inductive bias of attention structure, and the linear layer for embedding values is kept to maintain tokens' independence. 

The typical Transformer is built upon the idea of multi-head attention~\cite{dosovitskiy2020image} ($\rm MHA$), which constructs powerful representations at different positions from different subspaces. 
% That is defined as:
% \begin{equation}
% \label{mha}
% {\rm MHA}( Q, K, \textbf{V}) = {\rm Cat}( head_1, head_2, \cdots, head_h)W^O\ ,
% \end{equation}
% where $Q,\textbf{K},\textbf{V} \in \mathbb{R}^{n \times d_m}$ are input embedding matrices, $n$ is sequence length, $d_m$ is the embedding dimension, and h is the number of heads. Each head is defined as:
% \begin{equation}
% \label{att}
% \begin{split}
% head_j = {\rm Attention}_j( Q, \textbf{K}, \textbf{V})=\underbrace{{\rm softmax} \left[\frac{\hat{Q} \hat{\textbf{K}}^{\rm T}}{\sqrt{d_K}} \right]}_P \hat{\textbf{V}},\\
% \hat{Q}=QW_j^Q,
% \hat{\textbf{K}}=KW_j^\textbf{K},
% \hat{\textbf{V}}=VW_j^\textbf{V},
% \end{split}
% \end{equation}
% where $\hat{Q},\hat{\textbf{K}},\hat{\textbf{V}}$ are query, key and value projected into the subspaces respectively, $W_j^Q,W_j^\textbf{K} \in \mathbb{R}^{d_m \times d_K},W_j^\textbf{V} \in \mathbb{R}^{d_m \times d_V}$ are learned matrices, and $d_K, d_V$ are the hidden dimensions of the projection subspaces. The Attention defined in Eq. (\ref{att}) refers to a context mapping matrix $P \in \mathbb{R}^{n \times n}$.
The standard attention is $O(n^2)$ in time and space complexity due to the requirement to multiply two $n \times d$ matrices. Actually computing the full attention map is expensive and not suitable for local modeling. This work meticulously modifies the Transformer encoder with local-recognition attention to fuse features by setting $k \times k$ local view around every query of $\textbf{Q}$. Specifically, $d=c/h$, $h$ is set as the number of parallel attention heads. Then $\rm LRA$ of each head is defined as:
\begin{equation}
\label{LRA}
\begin{split}
\hat{\bm{h}}_j&={\rm LRA}_j( \textbf{Q} , \textbf{K}, \textbf{V} ) =\hat{\textbf{P}} \hat{\textbf{V}}=\mathcal{G}\left(\hat{\textbf{Q}},\hat{\textbf{K}}\right)\hat{\textbf{V}} \quad ,\\
\hat{\textbf{Q}}&=\phi_q(\textbf{Q}),\; \hat{\textbf{K}}=\phi_k(\textbf{K}),\; \hat{\textbf{V}}=\textbf{V}\textbf{W}_j^v \quad ,
\end{split}
\end{equation}
where $\hat{\bm{h}}_j$ denotes the process pipeline of $j$-th head. $\phi_q,\phi_k$ denote the convolution embedding for $\textbf{Q}$ and $\textbf{K}$ respectively, $\textbf{W}_j^v$ is the matrix for $\textbf{V}$'s linear embedding, $\mathcal{G}$ is the proposed local attention function, and $\hat{\textbf{P}} \in \mathbb{R}^{n \times n}$ is the local attention weight map. Each element of $\hat{P}$ is defined as:
\begin{equation}
\begin{split}
\hat{p}_{\hat{i},\hat{j}} &= \frac{
e^{
s_{
\hat{i},\hat{j}
}}}
{\sum\limits_{
\hat{g}=1}^n{
e^{s_{\hat{i},\hat{g}}
}}}\quad ,   
\\
\hat{s}_{\hat{i},\hat{j}}&=
\begin{cases}
\sum\limits_{\hat{l}=1}^{d_m}{\hat{q}_{\hat{i},\hat{l}}\hat{k}_{\hat{j},\hat{l}}}&{\hat{j} \in \hat{\textbf{S}}_{\hat{i}}}\\
{-\infty}&{\hat{j} \notin \hat{\textbf{S}}_{\hat{i}}}
\end{cases}\quad,
% \hat{s}_{\hat{i},\hat{j}}&=\left\{
% \begin{aligned}
% \sum\limits_{\hat{l}=1}^{d_m}{
% \hat{q}_{\hat{i},\hat{l}}
% \hat{k}_{\hat{j},\hat{l}}}, &{ }\\
% \quad \hat{j} \notin \mathcal{S}_{\hat{i}}\\
% \end{aligned}
\end{split}
\end{equation}
where $\hat{\textbf{S}}_{\hat{i}}$ is the selected local scope refer to position $\hat{i}$ which is defined as Fig. \ref{head1}, and $\hat{s}_{\hat{i},\hat{j}}$, $\hat{p}_{\hat{i},\hat{j}}$ represent the element of score map and $\hat{P}$ respectively. So the multi-head local-recognition attention ($\rm MH$-$\rm LRA$) is defined as:
\begin{equation}
{\rm MH}\text{-}{\rm LRA}( \textbf{Q}, \textbf{K}, \textbf{V}) = {\rm Cat}( \hat{h}_1, \hat{h}_2, \cdots, \hat{h}_h)\textbf{W}^O \quad ,
\label{MHLRA}
\end{equation}
where $\rm Cat$ is concatenating tensors in channel dimension, and $\textbf{W}^O$ is the matrix to project features.

The proposed local-recognition attention is $O(n)$ in time and space complexity and has more inductive bias based on a local receptive field to reduce the distraction of redundant information, which indicates the LRA's efficiency and accuracy on local modeling. The presented Transformer encoder shown in Fig. \ref{zt} deploys $\rm LRA$ for powerful local information representation, and builds local-modeling to global-search information process mechanism with standard global attention~\cite{vaswani2017nips}.

% \begin{remark}
% As is shown in Fig. \ref{block}, different from general attention, e.g., HiFT~\cite{cao2021iccv}, LRA sets identity to feature map $\textbf{Q}$ instead of $\textbf{V}$. This structure can fuse both feature maps effectively.
% \end{remark}

% \begin{figure}[t]
%       \centering
%       \includegraphics{graphic/am.pdf}
%       \caption{The proposed fusion attention can focus on the local region and contribute to fuse more valid information accurately.}
%       \label{big}
% \end{figure}

% \begin{figure}[t]
%       \centering
%       \includegraphics{graphic/4block.pdf}
%       \caption{Local-recognition attention. $M$ represent the output fusion feature. In this work, ${\mathcal K},\textbf{V}$ are same feature map, which is fused with feature map $\textbf{Q}$.}
%       \label{block}
% \end{figure}

%%%%%%%%%%%%%%%%%%%%%%%%%%%%%%%%%%%%%%%%%%%%%%%%%%%%%%%%%%%%%%%%%%%%%%%%%%%%%%%%
\subsection{Local Element Correction Network}

To further enhance the Transformer's ability to model local details, the $\rm LEC$ is proposed to guide the network to focus on local details.

$\rm LEC$ adopts networks with different receptive fields to search and refine different low-dimensional details. Consequently  a detail-correction map is generated. Specifically, $\rm LEC$ is defined as:
\begin{equation}
\label{LEC}
\begin{split}
\textbf{T} = {\rm LEC}(\textbf{Q},\textbf{K})={\rm DIN}({\rm Proj}({\rm Cat}(\textbf{Q},\textbf{K})))\quad ,\\
\end{split}
\end{equation}
where $\textbf{T}$ is the detail-correction map, ${\rm Proj}$ layer is adopted to project features to consistent channels, and $\rm DIN$ is the detail-inquire net which is denoted as:
\begin{equation}
\label{DIN}
\begin{split}
{\rm DIN}(\textit{{\textbf{x}}})=\textit{{\textbf{x}}}+ {\rm Proj}({\rm Cat}({\rm EG_{\uppercase\expandafter{\romannumeral1}}}(\textit{{\textbf{x}}}),{\rm EG_{\uppercase\expandafter{\romannumeral2}}}(\textit{{\textbf{x}}})))\quad ,
\end{split}
\end{equation}
where ${\rm EG}_*$ denotes an element generator and $\textit{{\textbf{x}}}$ is the input feature maps.

\Remark
The branch network is proposed to further increase the local details awareness. For capturing multi-details this work deploys blocks in GoogleNet~\cite{szegedy2015going} with different kernel sizes and sets smaller output channel numbers to construct the different element generators.

\begin{figure}[t]
       \centering
       \includegraphics[width=0.48\textwidth]{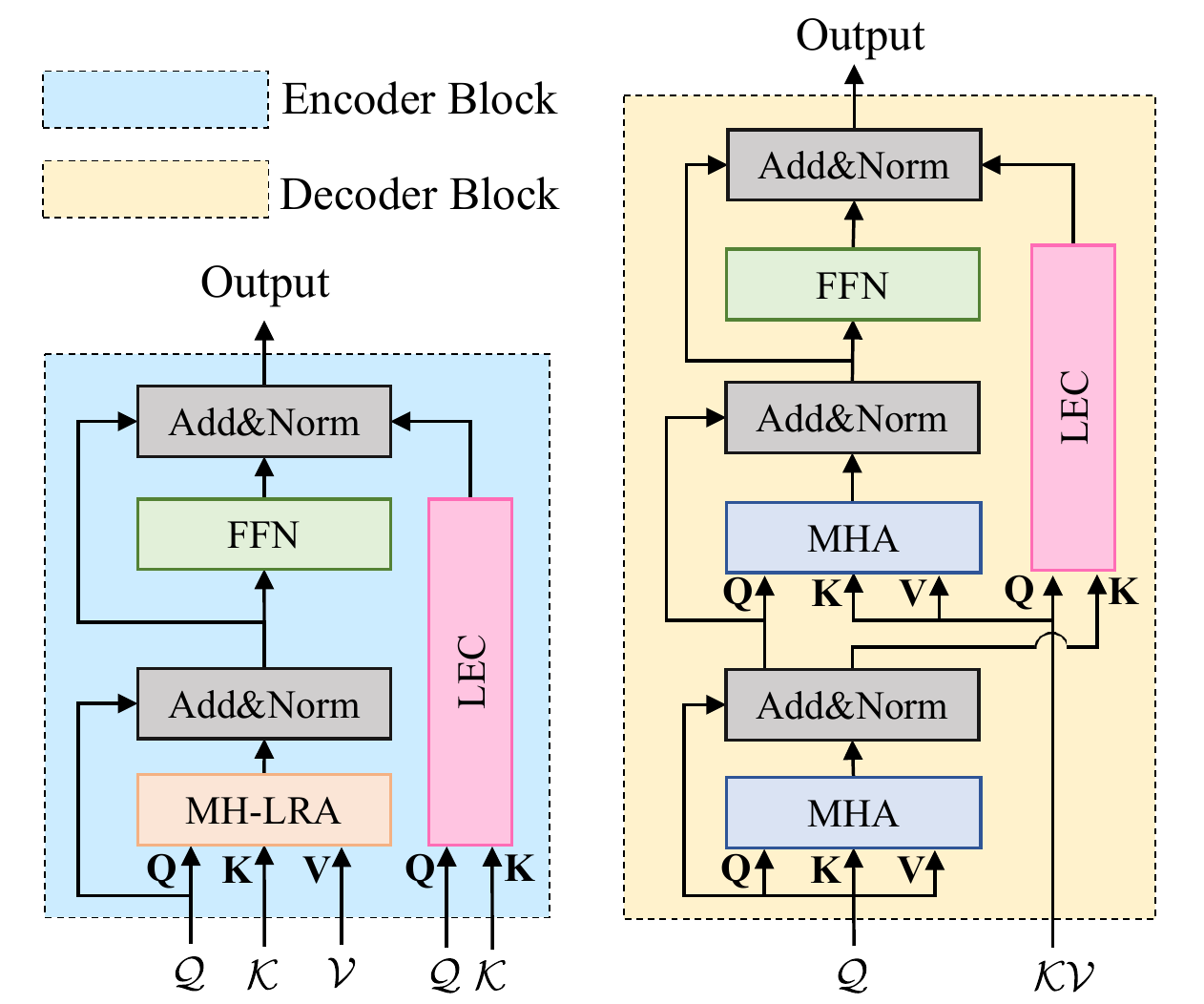}
       \vspace{-20pt}
       \caption{Left and right represent the proposed Transformer encoder and the standard Transformer decoder respectively. $\mathcal{Q}$, $\mathcal{K}$ and $\mathcal{V}$ represent the different input variables for a encoder/decoder module. Attributing to the combination of these two structures, LPAT gets the capability of efficient fusion and precise comparative search at the same time.}
       \label{trans}
       \vspace{-15pt}
\end{figure}
%%%%%%%%%%%%%%%%%%%%%%%%%%%%%%%%%%%%%%%%%%%%%%%%%%%%%%%%%%%%%%%%%%%%%%%%%%%%%%

%%%%%%%%%%%%%%%%%%%%%%%%%%%%%%%%%%%%%%%%%%%%%%%%%%%%%%%%%%%%%%%%%%%%%%%%%%%%%%%%
\subsection{Feature Processing with Local-Global Transformer}

Considering the strict requirements for model computational complexity under limited computing resources on aerial robots, LPAT adopts AlexNet~\cite{krizhevsky2012imagenet} to extract features. Both the template branch and the search branch adopt the first five CNN layers of AlexNet as the weight-shared backbone.
Specifically, this work denotes the template image and the search image as $\textbf{Z}$ and $\textbf{X}$ respectively. The $k$-th output of the template branch and the search branch are $\varphi_k(\textbf{Z})$ and $\varphi_k(\textbf{X})$ respectively. 

In order to compare the similarity between the template and search images, feature maps from $k$-th layer are input to depth-wise cross correlation layer (DWC), and the outputs are reshaped to $\textbf{M}_k \in \mathbb{R}^{WH \times C}$ as:
\begin{equation}
\textbf{M}_i = \mathcal{F}( \varphi_i(\textbf{Z}),\  \varphi_i(\textbf{X}))\ ,\ i=3,4,5 \  \quad ,
\end{equation}
where $C, W, H$ represent the channel, width, and height of the feature map respectively, and $\mathcal{F}$ denotes the depth-wise cross correlation layer. Then $\textbf{M}_3,\textbf{M}_4,\textbf{M}_5$ are input to the crafted Transformer module.

%%%%%%%%%%%%%%%%%%%%%%%%%%%%%%%%%%%%%%%%%%%%%%%%%%%%%%%%%%%%%%%%%%%%%%%%%%%%%%%%
\subsubsection{Transformer Encoder with Local Attention}

The proposed Transformer encoder adopts $\rm LRA$ for the efficiency of local modeling and $\rm LEC$ as a guide to catch local details. Specifically, as is shown in Fig. \ref{trans}, each layer of the encoder is constructed with $\rm MH$-$\rm LRA$ in Eq. (\ref{MHLRA}), $\rm LEC$, feed-forward network ($\rm FFN$), and layer normalization ($\rm Norm$).

% \begin{figure}[t]
%       \centering
%       \includegraphics[width=0.48\textwidth]{graphic/vis.pdf}
%       \vspace{-12pt}
%       \caption{Visualization of the proposed tracking pipeline. Before LPAT, disturbing noises in the search region may affect tracking performance, and LPAT can effectively reduce background interference and focus on the detailed representation of the tracked object to cope with extremely complex situation such as object rotation and deformation.}
%       \label{vis}
%       \vspace{-12pt}
% \end{figure}

Specifically the first encoder module treats $\textbf{M}_3$ as $\textbf{K}$ and $\textbf{V}$ and $\textbf{M}_4$ as $\textbf{Q}$ to get the first sequence of feature-fused tokens $\textbf{M}_{E1}$, which is denoted as:
\begin{equation}
\begin{split}
    \textbf{T}_{E1}&={\rm LEC}({\textbf{M}}_{3},{\textbf{M}}_{4})\quad ,\\
    \hat{\textbf{M}}_{E1}&={\rm Norm}({\rm MH}\text{-}{\rm LRA}(\textbf{M}_4,\textbf{M}_3,\textbf{M}_3)+\textbf{M}_4)\quad ,\\
    \textbf{\textbf{M}}_{E1}&={\rm Norm}(\textbf{T}_{E1}+{\rm Norm}({\rm FFN}(\hat{\textbf{\textbf{M}}}_{E1})+\hat{\textbf{\textbf{M}}}_{E1}))\quad ,
\end{split}
\end{equation}
where the $\hat{\textbf{\textbf{M}}}_{E1}$ is output of first normalization layer of first encoder module, and $\textbf{T}_{E1}$ is the first detail-correction map.

\Remark
Compared with the typical Transformer, the proposed local-recognition encoder has powerful representation of local details via $\rm LRA$ and $\rm LEC$, which keeps the tracker robust in object's appearance change.

Then, the second encoder layer inputs $M_{E1}$ as $\textbf{K}$ and $\textbf{V}$ and $M_5$ as $\textbf{Q}$, which is denoted as:
\begin{equation}
\begin{split}
    \textbf{T}_{E2}&={\rm LEC}({\textbf{M}}_{E1},{\textbf{M}}_{E1})\quad ,\\
    \hat{\textbf{M}}_{E2}&={\rm Norm}({\rm MH}\text{-}{\rm LRA}(\textbf{M}_5,\hat{\textbf{M}}_{E1},\hat{\textbf{M}}_{E1})+\textbf{M}_4)\quad ,\\
    \textbf{M}_{E2}&={\rm Norm}(\textbf{T}_{E2}+{\rm Norm}({\rm FFN}(\hat{\textbf{M}}_{E2})+\hat{\textbf{M}}_{E2}))\quad ,
\end{split}
\end{equation}
where $\hat{\textbf{M}}_{E2}$ is the output of first normalization layer in second encoder module, and  $\textbf{T}_{E2}$ is the second detail-correction map.

\Remark
In order to effectively fuse multi-scale information, LPAT uses two encoders to achieve a multi-level representation of deep features and shallow features to strengthen the modeling of local details.

% \begin{figure*}[thpb]
% \centering
% \includegraphics[width=1.0\textwidth]{graphic/6multi.pdf}
% \caption{PPs and SPs of the proposed method and other involved trackers on DTB70, UAV123@10fps and UAV123. These results indicate the large improvement of LPAT.}
% \label{sota}

% \end{figure*}

\begin{figure}[t]
      \centering
      \includegraphics[width=0.48\textwidth]{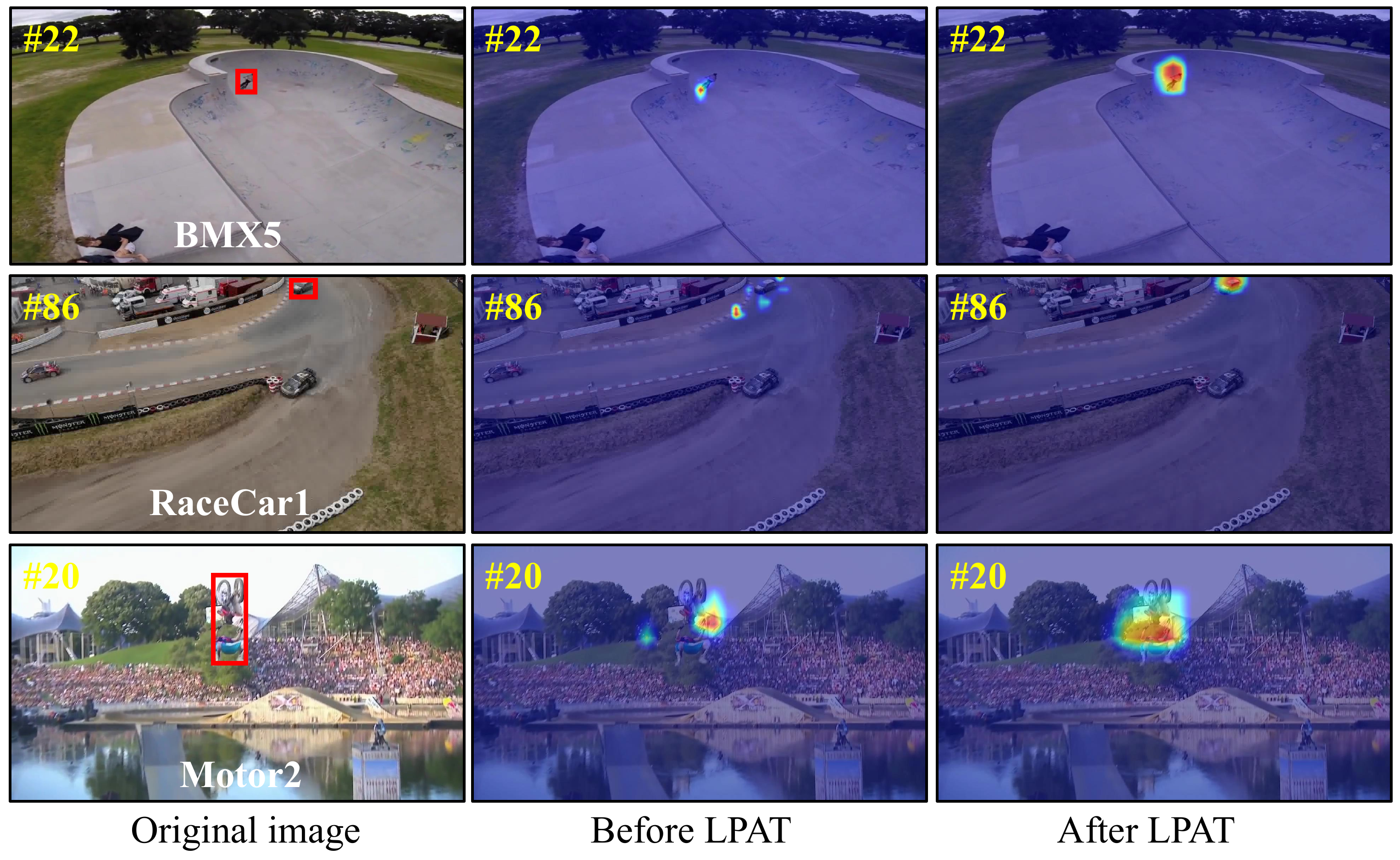}
      \vspace{-14pt}
      \caption{Visualization of the proposed tracking pipeline. All images are from benchmark DTB70~\cite{li2017visual}. Before LPAT, disturbing noises in the search region may affect tracking performance, and LPAT can effectively reduce background interference and focus on the detailed representation of the tracked object to cope with extremely complex situation such as object rotation and deformation.}
      \label{vis}
      \vspace{-15pt}
\end{figure}

\begin{figure*}[!t]
	\centering
		\includegraphics[width=0.325\linewidth]{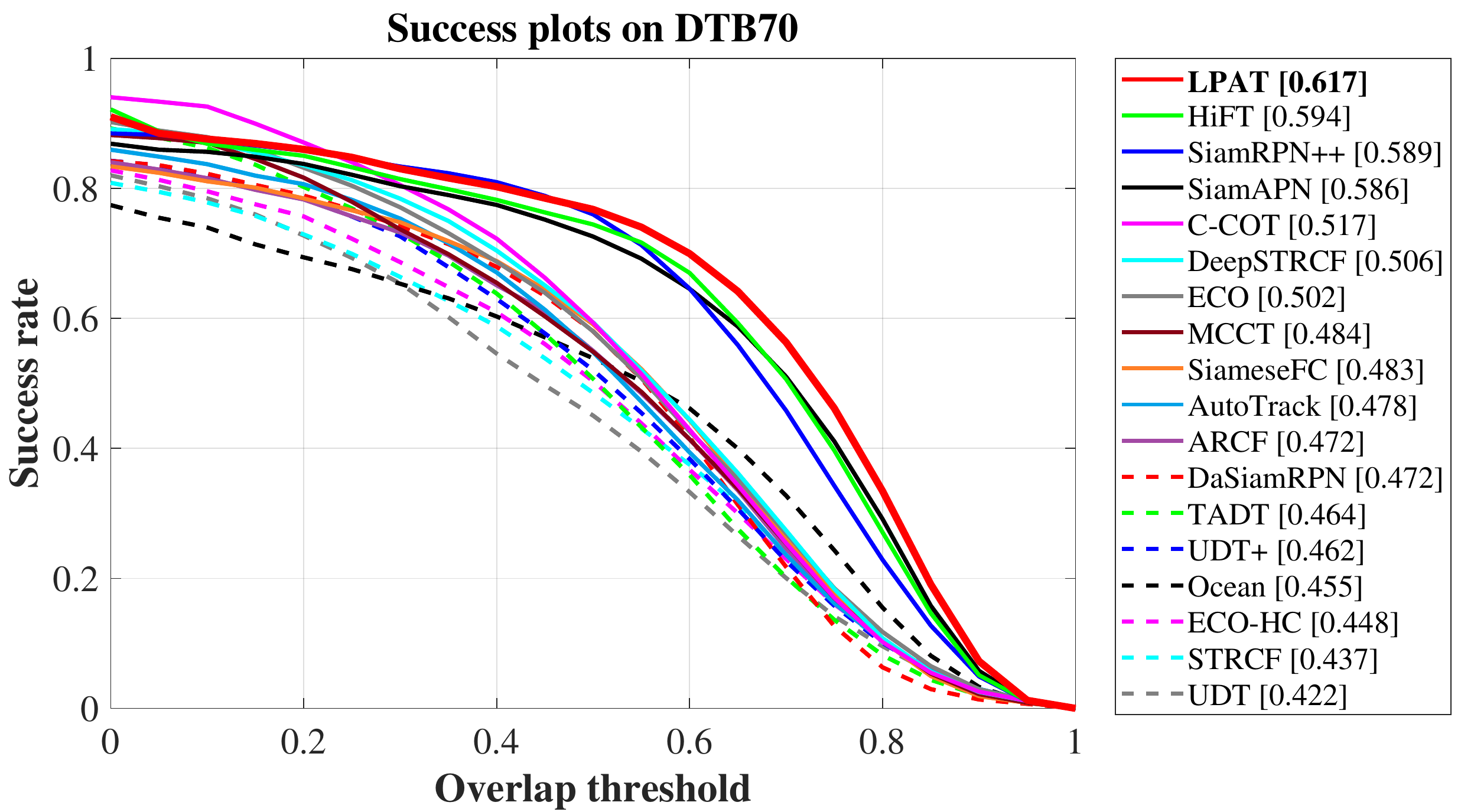}
		\includegraphics[width=0.325\linewidth]{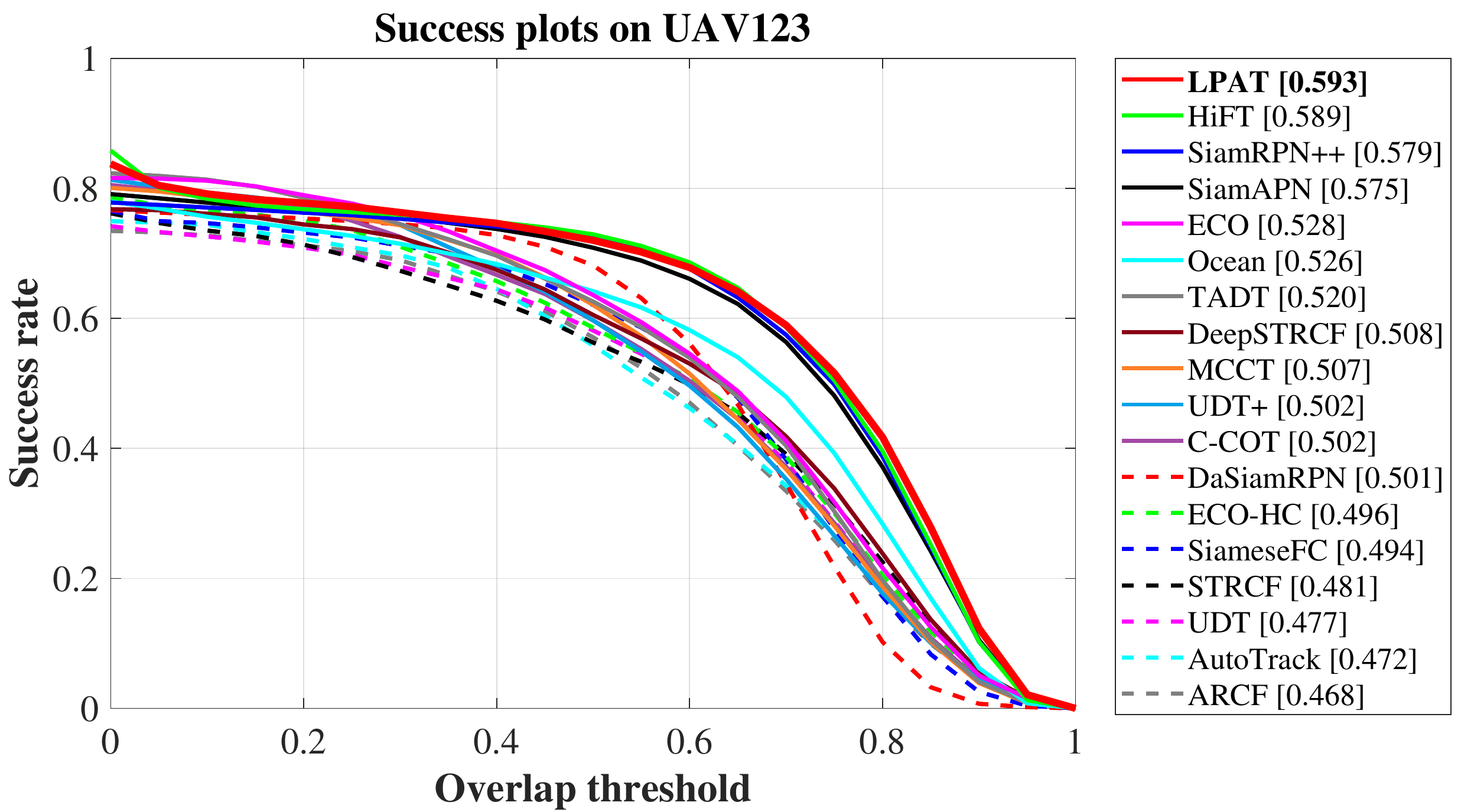}
		\includegraphics[width=0.325\linewidth]{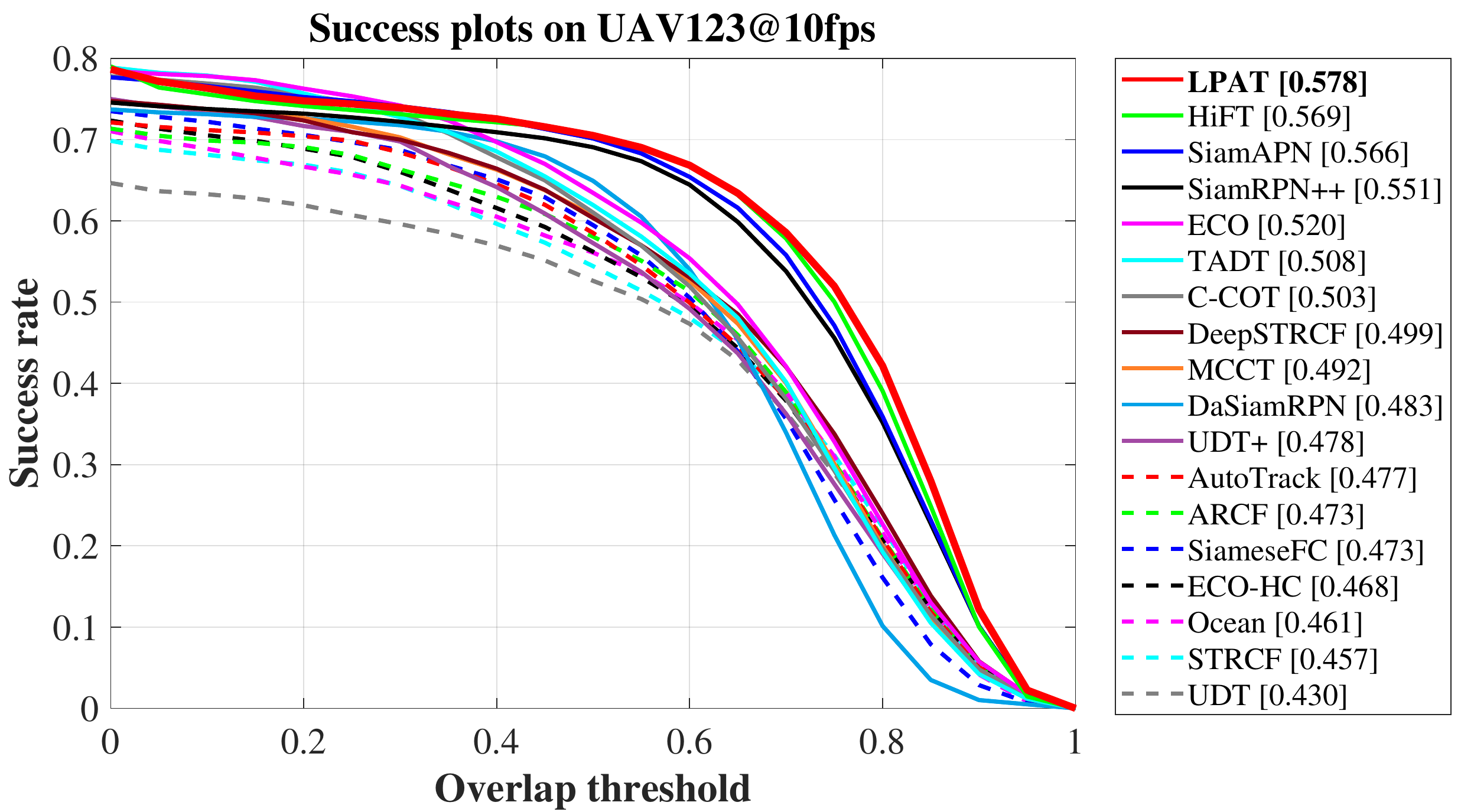}

	\includegraphics[width=0.325\linewidth]{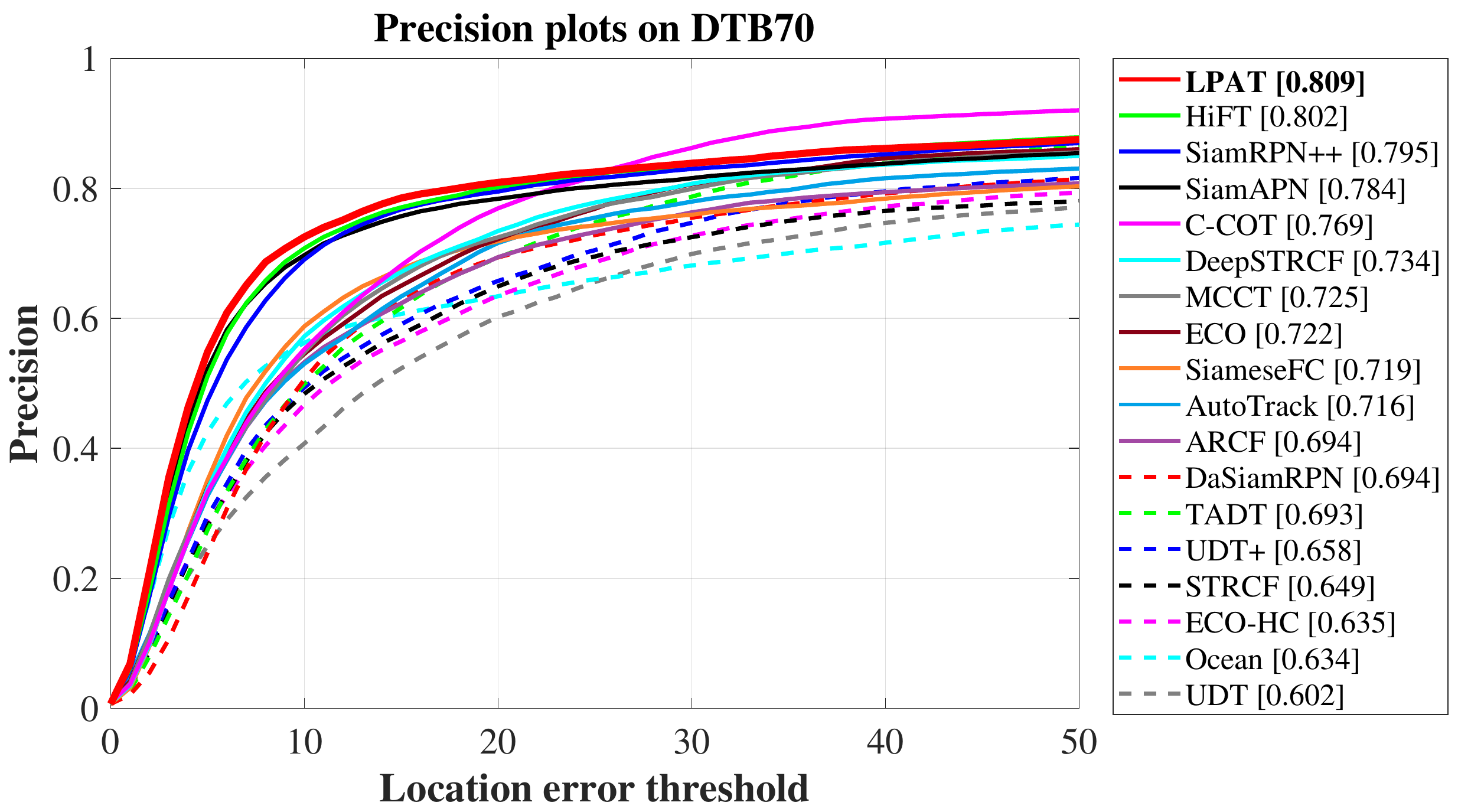}
		\includegraphics[width=0.325\linewidth]{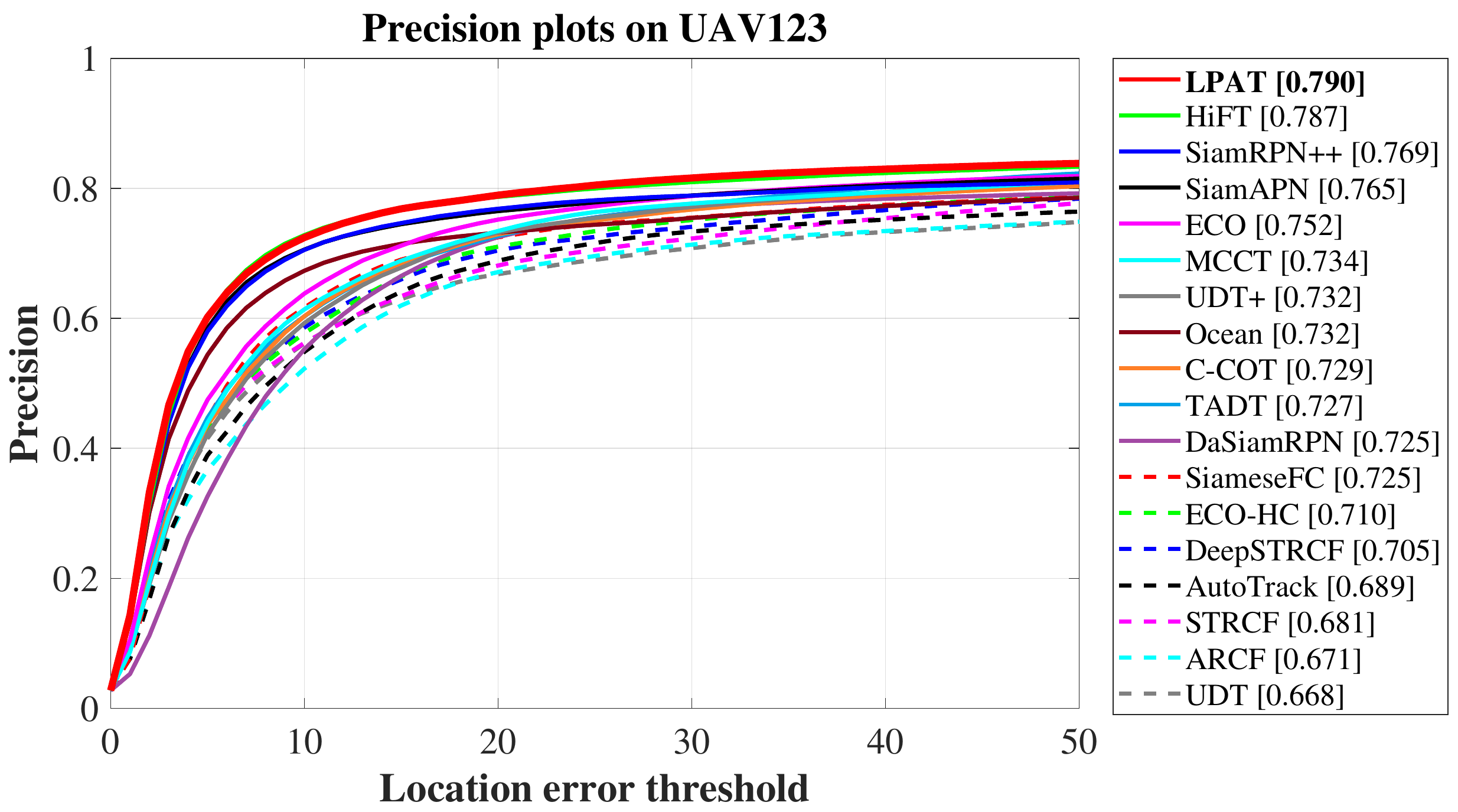}
		\includegraphics[width=0.325\linewidth]{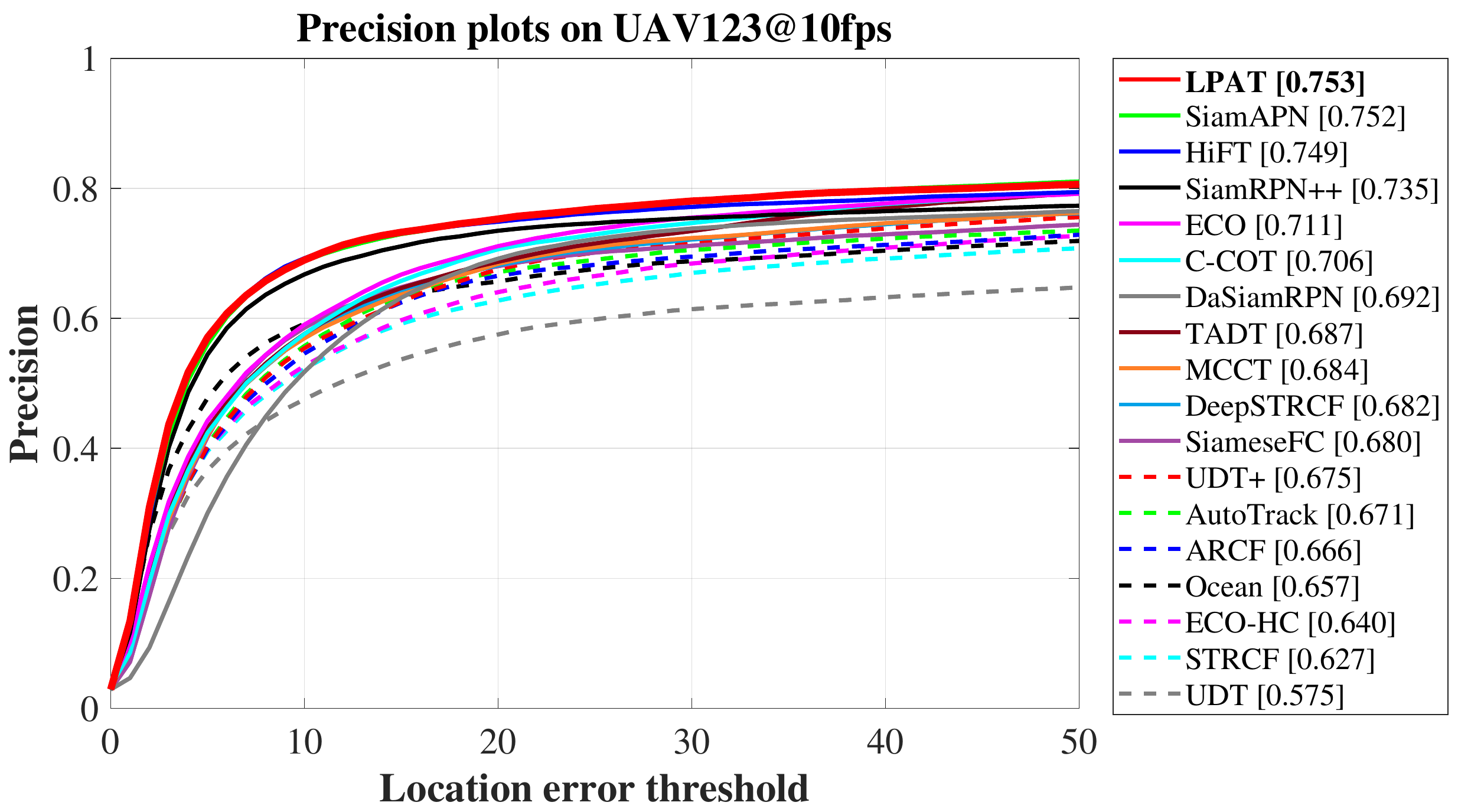}

	\caption
	{
		PPs and SPs of the proposed method and other involved SOTA trackers on authoritative benchmarks, \textit{i.e.}, DTB70, UAV123, and UAV123@10fps. The results indicate the large improvement of LPAT.
	}
	\label{sota}
	\vspace{-19pt}
	%	\setlength{\abovecaptionskip}{-5pt} 
	%	\vspace{-10pt}
\end{figure*}

%%%%%%%%%%%%%%%%%%%%%%%%%%%%%%%%%%%%%%%%%%%%%%%%%%%%%%%%%%%%%%%%%%%%%%%%%%%%%%%%
\subsubsection{Transformer Decoder with Global Attention}

The proposed Transformer decoder adopts global attention for global search. Specifically, as is shown in Fig. \ref{trans} each layer of the encoder is constructed with $\rm MHA$, $\rm LEC$, $\rm FFN$, and $\rm Norm$.

The proposed method adopts $\rm DIN$ to preset the initial tokens of target queries for its rich semantic information, and then treats the final output tokens of encoder, $\hat{\textbf{M}}_{E2}$, as $\textbf{K},\textbf{V}$. A $\rm LEC$ layer is also used to guide global-search Transformer to catch local details. The decoder' output sequences ($\textbf{M}_{D}$) is denoted as:
\begin{equation}
\begin{split}
   % M_t&={\rm DIN}(M_5),\\
    % \textbf{Q}_D&={\rm DIN}(\textbf{M}_5)\quad ,\\
    \hat{\textbf{Q}}_{D}&={\rm Norm}({\rm MHSA}({\rm DIN}(\textbf{M}_5))\quad ,\\
    %{T}_{D}&=\text{LEC}(\hat{Q}_{D},\hat{M}_{E2}),\\
    \textbf{T}_{D}&={\rm LEC}(\hat{\textbf{Q}}_{D},\hat{\textbf{M}}_{E2})\quad ,\\
    \hat{\textbf{M}}_{D}&={\rm Norm}({\rm MHA}(\hat{\textbf{Q}}_{D},\hat{\textbf{M}}_{E2},\hat{\textbf{M}}_{E2})+\hat{\textbf{Q}}_{D})\quad ,\\
    \textbf{M}_{D}&={\rm Norm}(\textbf{T}_{D}+{\rm Norm}({\rm FFN}(\hat{\textbf{M}}_{D})+\hat{\textbf{M}}_{D}))\quad,\\
\end{split}
\end{equation}
where $\hat{\textbf{Q}}_{D}$, $\textbf{T}_{D}$, $\hat{\textbf{M}}_{D}$ and  is the output of the multi-head self-attention layer ($\textbf{MHSA}$), the cross attention layer and $\rm LEC$ respectively. The visualization of the proposed track pipeline is reported in Fig. \ref{vis}. Sophisticated local modeling and efficient global search improve the LPAT's robustness.
% The second decoder is denoted as:
% \begin{equation}
% \begin{split}
%     \hat{Q}_{D}&={\rm Norm}({\rm MHA}(M_5,M_5,M_5),\\
%     \hat{M}_{D}&={\rm Norm}({\rm MHA}(\hat{Q}_{D},\hat{M}_{E2},\hat{M}_{E2})+\hat{Q}_{D}),\\
%     M_{D}&={\rm Norm}({\rm FFN}(\hat{M}_{D})+\hat{M}_{D}),
% \end{split}
% \end{equation}
% where $\hat{Q}_{D}$, $\hat{M}_{D}$, $M_{D}$ is the output of self-attention layer, cross attention layer, and FFN layer respectively. 

\Remark
The proposed local-global mechanism is built by a cross-module construction method, instead of in-layer feature procession~\cite{chu2021twins}, which makes LPAT more robust.

%%%%%%%%%%%%%%%%%%%%%%%%%%%%%%%%%%%%%%%%%%%%%%%%%%%%%%%%%%%%%%%%%%%%%%%%%%%%%%%%

% if have a single appendix:
%\appendix[Proof of the ZonkLRA Equations]
% or
%\appendix  % for no appendix heading
% do not use \section anymore after \appendix, only \section*
% is possibly needed

% use appendices with more than one appendix
% then use \section to start each appendix
% you must declare a \section before using any
% \subsection or using \label (\appendices by itself
% starts a section numbered zero.)
%

%%%%%%%%%%%%%%%%%%%%%%%%%%%%%%%%%%%%%%%%%%%%%%%%%%%%%%%%%%%%%%%%%%%%%%%%%%%%%%%%
\subsection{Classification \& Regression}
This work adopts two classification branches and one regression branch~\cite{cao2021hift} which are all constructed by CNN. The cross-entropy loss ($\mathcal{L}_{cls1}$) is applied to the first classification branch for classification via ground truth area. The binary cross-entropy loss ($\mathcal{L}_{cls2}$) is applied to the other branch to explore area near the ground truth center point. The regression branch is supervised by IOU loss ($\mathcal{L}_{reg}$). Therefore, setting $\lambda_*$ as the hyper-parameter of each different loss, this work determines the overall loss function as:
\begin{equation}
\begin{split}
    \mathcal{L}=\lambda_1\mathcal{L}_{cls1}+\lambda_2\mathcal{L}_{cls2}+\lambda_3\mathcal{L}_{reg}\quad .
\end{split}
\end{equation}

%%%%%%%%%%%%%%%%%%%%%%%%%%%%%%%%%%%%%%%%%%%%%%%%%%%%%%%%%%%%%%%%%%%%%%%%%%%%%%%

%%%%%%%%%%%%%%%%%%%%%%%%%%%%%%%%%%%%%%%%%%%%%%%%%%%%%%%%%%%%%%%%%%%%%%%%%%%%%%%%
\section{Experiments}\label{exper}
%%%%%%%%%%%%%%%%%%%%%%%%%%%%%%%%%%%%%%%%%%%%%%%%%%%%%%%%%%%%%%%%%%%%%%%%%%%%%%%%
\subsection{Experimental Setup}

\noindent\textbf{Evaluation Methodology:}
In this work, an one-pass evaluation protocol is deployed to evaluate all trackers' performances via two metrics of success rate and precision ~\cite{mueller2016benchmark}. The percentage of frames where the center location error (CLE) is within a given threshold, is demonstrated by the precision. The percentage of frames where the overlap exceeds a given threshold, is visualized via the success rate.

\noindent\textbf{Implementation Details:}
Image pairs extracted from ImageNet VID~\cite{russakovsky2015imagenet}, COCO~\cite{lin2014microsoft}, GOT-10K~\cite{huang2019got} and Youtube-BB~\cite{real2017youtube} are used to train the tracker. The sizes of the search region patch and template patch are $287 \times 287$ and $127 \times 127$, respectively.
The proposed framework adopts the modified AlexNet~\cite{krizhevsky2012imagenet} pretrained with ImageNet~\cite{russakovsky2015imagenet} as described in SiamRPN~\cite{li2018high}. Specifically, the whole network is totally trained for 70 epochs and this work fine-tunes the last three layers of the AlexNet in the last 60 epochs. The learning rate of the back-end networks is initiated as $5\times10^{-4}$ and decreases in the log space from $10^{-2}$ to $10^{-4}$. The training process adopts the stochastic gradient descent (SGD). Additionally, the decay settings of batch size, momentum, and weight are 220, 0.9, and $10^{-4}$, respectively. The hardware configuration of the PC deployed to train the network is two NVIDIA TITAN RTX GPUs, an Intel i9-9920X CPU and 32GB RAM. All trackers in this section are evaluated with one NVIDIA TITAN RTX GPU. Our source code and some demo videos are available at \url{https://github.com/vision4robotics/LPAT}.

\subsection{Overall Performance}

The proposed methods are compared with the state-of-the-art trackers including HiFT~\cite{cao2021hift}, SiamRPN++~\cite{Li2019CVPR}, SiamAPN~\cite{fu2021onboard}, DaSiamRPN~\cite{zhu2018distractor}, C-COT~\cite{danelljan2016beyond}, DeepSTRCF~\cite{li2018learning}, ECO~\cite{danelljan2017eco}, ECO-HC~\cite{danelljan2017eco}, Ocean~\cite{zhang2020ocean}, UDT+~\cite{wang2019unsupervised}, SiameseFC~\cite{Bertinetto2016ECCVW}, ARCF~\cite{huang2019learning}, AutoTrack~\cite{li2020autotrack}, TADT~\cite{li2019target} and MCCT\cite{wang2018multi}. For fairness, a same backbone, $i.e.$, AlexNet pretrained on ImageNet~\cite{russakovsky2015imagenet} is adopted in all involved Siamese network-based trackers. The detailed comparison results on UAV123~\cite{mueller2016benchmark}, UAV123@10fps~\cite{mueller2016benchmark}, and DTB70~\cite{li2017visual} benchmarks are reported.

\begin{table}[t]
\centering

\caption{Attribute-based success evaluation on DTB70. \textbf{\textcolor{red}{red}}, \textbf{\textcolor{green}{green}}, and \textbf{\textcolor{blue}{blue}} colors are adopted to highlight he best three performances in the table.}
\vspace{-8pt}

\label{attr}
\begin{tabular}{l|cccccc}
\toprule
Attr. & Def. & FCM & IPR & MB & OPR & SV\\
\hline
    UDT   & 0.407  & 0.434  & 0.389  & 0.385  & 0.326  & 0.464  \\
    STRCF & 0.390  & 0.467  & 0.393  & 0.447  & 0.257  & 0.417  \\
    ECO-HC & 0.389  & 0.464  & 0.401  & 0.426  & 0.311  & 0.429  \\
    Ocean & 0.371  & 0.492  & 0.412  & 0.409  & 0.394  & 0.355  \\
    UDT+  & 0.386  & 0.476  & 0.420  & 0.447  & 0.289  & 0.425  \\
    TADT  & 0.449  & 0.466  & 0.433  & 0.450  & 0.407  & 0.454  \\
    DaSiamRPN & 0.493  & 0.457  & 0.435  & 0.404  & 0.416  & 0.531  \\
    ARCF  & 0.426  & 0.496  & 0.429  & 0.453  & 0.321  & 0.487  \\
    AutoTrack & 0.452  & 0.496  & 0.454  & 0.468  & 0.343  & 0.493  \\
    SiameseFC & 0.467  & 0.487  & 0.447  & 0.453  & 0.367  & 0.483  \\
    MCCT  & 0.466  & 0.494  & 0.448  & 0.478  & 0.364  & 0.510  \\
    ECO   & 0.424  & 0.514  & 0.451  & 0.496  & 0.314  & 0.481  \\
    DeepSTRCF & 0.496  & 0.516  & 0.475  & 0.477  & 0.374  & 0.518  \\
    C-COT & 0.430  & 0.551  & 0.477  & 0.522  & 0.356  & 0.479  \\
    SiamRPN++ & 0.579  & 0.587  & 0.566  & \textbf{\textcolor{blue}{0.531}}  & 0.484  & 0.637  \\
    SiamAPN & \textbf{\textcolor{blue}{0.616}}  & \textbf{\textcolor{blue}{0.599}}  & \textbf{\textcolor{blue}{0.572}}  & 0.525  & \textbf{\textcolor{blue}{0.518}}  & \textbf{\textcolor{green}{0.667}}  \\
    HiFT  & \textbf{\textcolor{green}{0.626}}  & \textbf{\textcolor{green}{0.611}}  & \textbf{\textcolor{green}{0.613}}  & \textbf{\textcolor{green}{0.589}}  & \textbf{\textcolor{green}{0.546}}  & \textbf{\textcolor{blue}{0.649}}  \\
\hline
\textbf{LPAT} & \textbf{\textcolor{red}{0.640}} & \textbf{\textcolor{red}{0.620}} & \textbf{\textcolor{red}{0.623}} & \textbf{\textcolor{red}{0.595}} & \textbf{\textcolor{red}{0.550}} & \textbf{\textcolor{red}{0.670}}\\
\hline
\end{tabular}
\vspace{-12pt}
\end{table}

\begin{table}[b]
    \centering
    \setlength{\belowcaptionskip}{7pt}
    \vspace{-12pt}
    \caption{Ablation Study for the proposed method. LPAT achieves the great performance with fully method.}
    \vspace{-8pt}

    \resizebox{.95\columnwidth}{!}{

    \scalebox{1.0}{
    \begin{tabular}{ccccc}
         \toprule
         \multicolumn{1}{c}{Methods}&\multicolumn{1}{c}{Precision}&\multicolumn{1}{c}{$\Delta_{pre} (\%)$}&
         \multicolumn{1}{c}{Success}&\multicolumn{1}{c}{$\Delta_{suc} (\%)$}\\
         \midrule
         \multicolumn{1}{c}{Baseline}&\multicolumn{1}{c}{0.751} & $-$ & 0.562 & $-$ \\
         \multicolumn{1}{c}{$\rm LRA$}&\multicolumn{1}{c}{0.799} & 6.39 & 0.587 & 4.45 \\
         \multicolumn{1}{c}{$\rm LRA$+$\rm LEC$}&\multicolumn{1}{c}{0.792} & 5.46 & 0.602 & 7.12 \\
         \midrule
         \multicolumn{1}{c}{\textbf{$\rm LRA$+$\rm LEC$+LG (LPAT)}}&\multicolumn{1}{c}{\textbf{0.809}} & \textbf{7.19} & \textbf{0.617} & \textbf{9.78}\\
         \bottomrule
    \end{tabular}
    }}
    \label{tab:ablat}%
\end{table}

% \begin{figure*}[t]
% \centering
% \includegraphics[width=16cm]{rwt.PNG}
% \caption{Multi-compare. Compared with other trackers, the proposed method combines excellent speed and accuracy and proves its practicality on UAV tracking.}
% \label{attr}
% \end{figure*}
%%%%%%%%%%%%%%%%%%%%%%%%%%%%%%%%%%%%%%%%%%%%%%%%%%%%%%%%%%%%%%%%%%%%%%%%%%%%%%%%
\noindent\textbf{DTB70~\cite{li2017visual}:} 
 Containing 70 challenging UAV sequences with various extremely complex scenarios, DTB70 is a typical benchmark to evaluate the tracks' performance. Specifically, the numerous challenges in each sequence include deformation, in-plane rotation, scale variation, \textit{etc.} As presented in the first column of Fig. \ref{sota}, the proposed tracker gains the best success rate (\textbf{0.617}), outperforming the second-best HiFT (0.594) and the third-best SiamRPN++ (0.589) by \textbf{3.87}\% and \textbf{4.75}\%. Meanwhile, in the precision, LPAT also surpasses others with the best precision \textbf{0.809}, while the second-best HiFT and the third-best SiamRPN++ gain 0.802 and 0.795 respectively.

%%%%%%%%%%%%%%%%%%%%%%%%%%%%%%%%%%%%%%%%%%%%%%%%%%%%%%%%%%%%%%%%%%%%%%%%%%%%%%%%
\noindent\textbf{UAV123@30fps~\cite{mueller2016benchmark}:} 
Consisting of 123 sequences with more than 112K frames, the benchmark is adopted to evaluate the performances in a low-altitude aerial perspective. In this benchmark, trackers are facing the numerous challenging scenarios in aerial tracking, \textit{e.g.} scale variation and fast motion. As illustrated in the third column of Fig.~\ref{sota}, LPAT outperforms others in success rate and achieves the best performance with \textbf{0.593} success rate and \textbf{0.790} precision.

\noindent\textbf{UAV123@10fps~\cite{mueller2016benchmark}:} 
Sequences in this benchmark are downsampled from those recorded in 30 fps. Consequently, the attribute of object motion in UAV123@10fps is more severe and challenging, which is reported in the third column of Fig. \ref{sota}. LPAT outperforms other trackers with \textbf{0.578} success rate and \textbf{0.753} precision. 

The higher performances on the above benchmarks indicate the accuracy and robustness of LPAT, which shows the powerful local modeling and effective global searching.

% \begin{table}[t]  
% \centering  
% \caption{Ablation result.}  
% \begin{tabular}{lllll}
% 1 & 1 & 1 & 1 & 1\\
% 1 &   &   &   &   \\
% 1 & 1 & 1 & 1 & 1\\
% 1 &   &   &   &   \\
% 1 & 1 & 1 & 1 & 1\\
% 1 &   &   &   &   \\
% 1 & 1 & 1 & 1 & 1
% \end{tabular}
% \end{table}

\subsection{Attribute-based Comparison}

The robustness of LPAT under extremely complex challenges is evaluated via attribute-based comparisons in DTB70~\cite{li2017visual}. The success rate is adopted to describe the accuracy and robustness. Results of success rates on several extremely complex challenges are reported in TABLE \ref{attr}, including deformation (Def.), fast camera motion (FCM), in-plane rotation (IPR), motion blur (MB), out-of-plane rotation (OPR) and scale variation (SV). The proposed tracker achieves the best scores in most scenarios. Especially in in-plane rotation and deformation, the powerful local details perception-aware framework with meticulous local modeling increases the performance of global search via effective perception for object's multiple variation. 
\begin{table*}[t]  
\centering  
\caption{Contrast with deeper backbone-based trackers in DTB70. The proposed method's best performance benefits from the efficient Transformer framework. \textbf{\textcolor{red}{red}}, \textbf{\textcolor{green}{green}}, and \textbf{\textcolor{blue}{blue}} colors are adopted to highlight the best three performances respectively.}
\vspace{-7pt}
\label{DB}
\begin{tabular}{lcccccccccc}
\toprule
% Attributes & Aspect ratio variation & Background cluctter & Fast camera motion & In-plane rotation & Motion blur & Occllusion & Out-of-plane & Out-of-view & Scale variation & SimiLRA objects around\\
% Attributes & Aspect ratio variation & Background cluctter & Fast camera motion & In-plane rotation & Occllusion & Out-of-view & Scale variation\\
Trackers & SiamRPN++~\cite{Li2019CVPR}& SiamDW\_RPN~\cite{SiamDW_2019_CVPR} & SiamGAT\cite{guo2021graph} & SiamMASK~\cite{wang2019fast} & SiamCAR~\cite{guo2020siamcar} &  SiamRPN++~\cite{Li2019CVPR} & \textbf{LPAT(ours)}\\
\midrule
% trackers & Aspect ratio variRPation & Background cluctter & Fast camera Out-of-view & Scale variation & SimiLRA objects around\\
Backbone & MobileNetV2~\cite{Sandler2018MobileNetV2} & ResNet-22~\cite{he2016deep} & GoogleNet~\cite{szegedy2015going} & ResNet-50~\cite{he2016deep} & ResNet-50~\cite{he2016deep} & ResNet-50~\cite{he2016deep} & AlexNet~\cite{krizhevsky2012imagenet}\\
Success & 0.593 & 0.453 & 0.583 & 0.571 & \textbf{\textcolor{blue}{0.597}} & \textbf{\textcolor{green}{0.615}} & \textbf{\textcolor{red}{0.617}}\\
Precision & 0.785& 0.709 & 0.752 & 0.769 & \textbf{\textcolor{blue}{0.801}} & \textbf{\textcolor{green}{0.802}} & \textbf{\textcolor{red}{0.809}}\\
% FPS &   &   &  &  &  &  &  \textbf{\textcolor{green}{}} & \textbf{\textcolor{red}{}}\\
\bottomrule
\end{tabular}
\vspace{-18pt}
\end{table*}

\begin{figure}[t]
       \centering
       \includegraphics{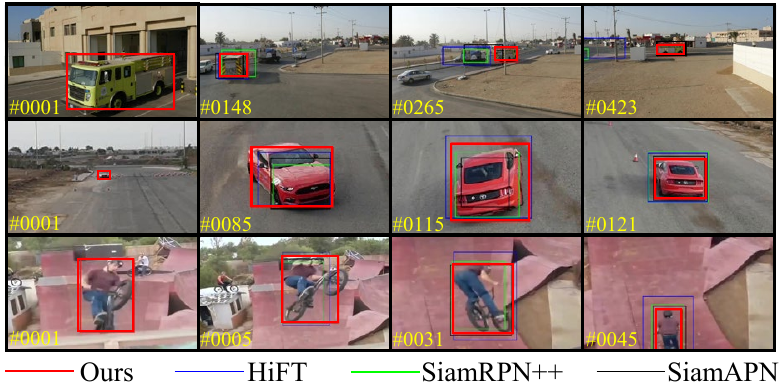}
       \vspace{-8pt}
       \caption{Screenshots of \textit{truck1} from UAV123@30fps, \textit{car16\_1} from UAV123@10fps, and \textit{BXM3} from DTB70. The proposed LPAT performs excellently in extremely complex aerial object tracking scenario. }
       \label{comp}
       \vspace{-19pt}
\end{figure}

%%%%%%%%%%%%%%%%%%%%%%%%%%%%%%%%%%%%%%%%%%%%%%%%%%%%%%%%%%%%%%%%%%%%%%%%%%%%%%%%
\subsection{Ablation Study}

The impact of each block and local-to-global mechanism proposed in this work is analyzed. Baseline is considered as the feature extraction, feature process with global attention-based typical Transformer and the same prediction head. $\rm LRA$ denotes adopting local-recognition attention to replace attention blocks in encoder and processes information by local to local mechanism in baseline. $\rm LEC$ denotes adopting local element correction network for all Transformer blocks. LG denotes adopting local to global information processing method. The success and precision are reported in TABLE \ref{tab:ablat}. Compared with Baseline, LPAT is indicated that the proposed blocks, $\rm LRA$ and $\rm LEC$ both can effectively increase the accuracy of the Baseline, and the proposed local-modeling to global-search information processing mechanism is also indispensable for LPAT.

\subsection{Comparison to Trackers with Deeper Backbone}\label{D}

To verify the efficiency, the presented method is compared in DTB70~\cite{li2017visual} with other trackers with deeper backbones, \textit{i.e.}, SiamRPN++ (ResNet-50)~\cite{Li2019CVPR}, SiamRPN++ (MobileNetV2)~\cite{Li2019CVPR}, SiamMask (ResNet-50)~\cite{wang2019fast}, SiamCAR (ResNet-50)~\cite{guo2020siamcar}, SiamDW\_RPN (ResNet-22)~\cite{SiamDW_2019_CVPR} and SiamGAT (GoogleNet)~\cite{guo2021graph}. The results are reported in TABLE \ref{DB}. LPAT performs the best on both success rate and precision with AlexNet which only has five layers. This comparison indicates the superior efficiency and performance of the proposed method.

\subsection{Qualitative Evaluations}\label{Qualitative}
Some qualitative evaluations are released in Fig. \ref{comp}. Compared with several state-of-the-art trackers~\cite{cao2021hift,fu2021onboard,Li2019CVPR}, LPAT is able to more appropriately predict bounding box in multiple extremely complex challenges, \textit{e.g.}, partial occlusion, in-plane rotation, fast motion, scale variation and low resolution. It is clearly illustrated that the proposed tracker designed for object tracking maintains superior accuracy and robustness via powerful local perception.

\begin{figure}[t]
\centering
\includegraphics[width=8.5cm]{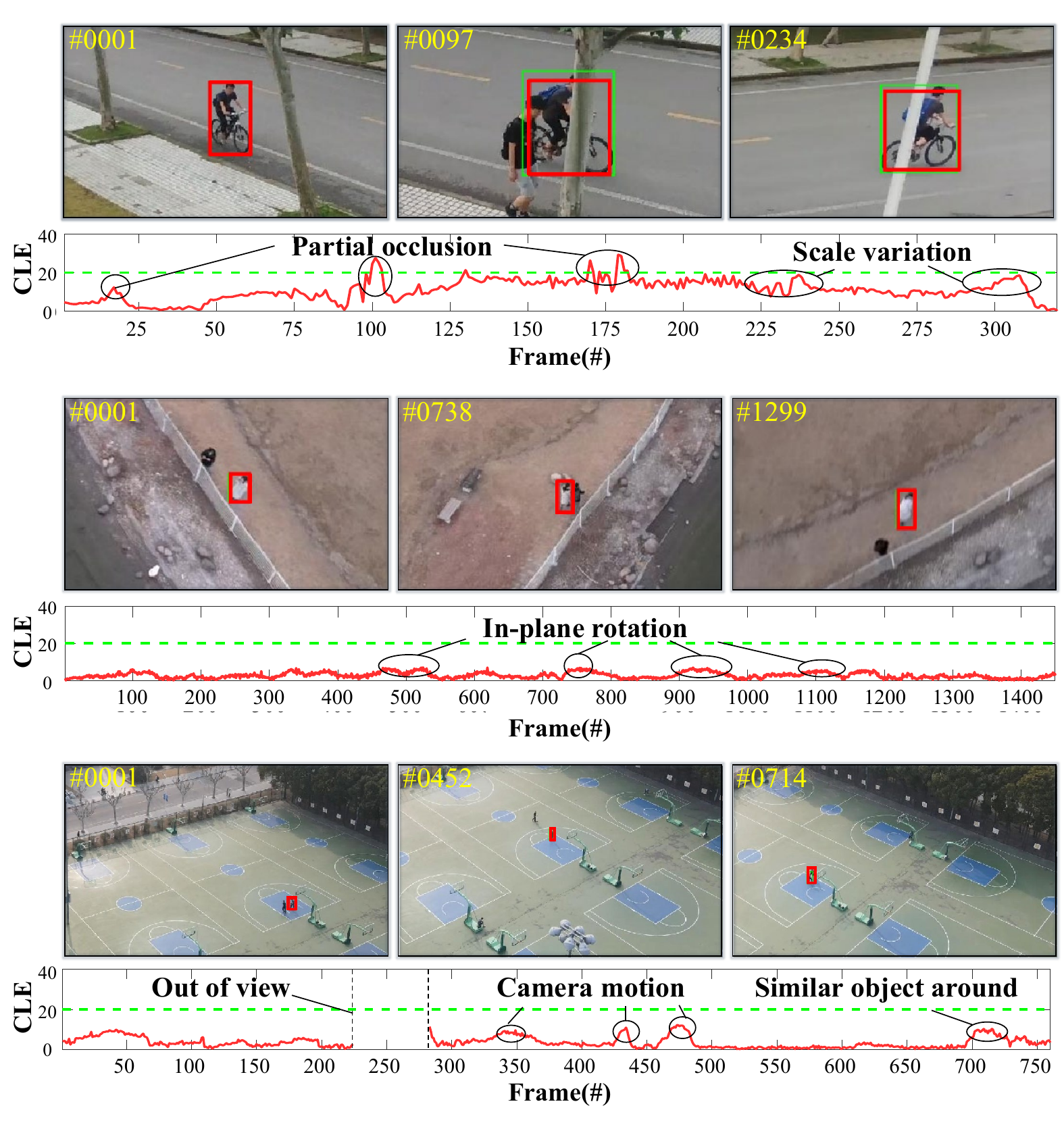}
\vspace{-14pt}
\caption{Real-world tests. The proposed tracker is tested on a common UAV platform with typical embedded processor, \textit{i.e.}, NVIDIA Jetson AGX Xavier. The tracking results and ground truth are marked with \textbf{\textcolor{red}{red}} and \textbf{\textcolor{green}{green}} box respectively.}
\label{RealT}
\vspace{-18pt}
\end{figure}

%%%%%%%%%%%%%%%%%%%%%%%%%%%%%%%%%%%%%%%%%%%%%%%%%%%%%%%%%%%%%%%%%%%%%%%%%%%%%%%%
\section{Real-world Tests}
%从评估中得知已经获得最佳的性能，接下来进一步验证在实际环境中的performence
Besides the superior performance of LPAT which is shown in Sec.~\ref{exper}, this section reports the LPAT's capability in real-world scene. The presented tracker is deployed on a common UAV platform with typical embedded processor, \textit{i.e.}, NVIDIA Jetson AGX Xavier. In Fig. \ref{RealT}, the tests' results are presented. It is noted that the tests face many challenges in complex scenes, especially scale variation, partial occlusion (the first row), in-plane rotation, low-resolution (the second row), out of view, camera motion and small object (the last row). The proposed tracker contrusted with the meticulous local-to-global Transformer framework achieves considerable results in complex actual environments and maintains about 21FPS in the tests. The promising performances indicate the LPAT's strong robustness in different challenges and the efficiency in aerial application scenarios.

\section{Conclusions}

This work presents a unified framework working in a local-global mechanism to fuse features and directly input to a prediction head for the final tracking result. $\rm LRA$ not only reduces the consumption of computing resources but also suppresses the introduction of interference information. In another parallel pipeline, $\rm LEC$ also improves the Transformer modules' local details representation abilities. Attributing to the efficiency of local-global mechanism, the presented tracker achieves promising performances in different benchmarks and real-world tests with competitive speed. It is convinced that more efficient blocks with the local inductive bias and more delicate local-to-global mechanism will further improve Transformer-based aerial object tracking.
   % This command serves to balance the column lengths
                                  % on the last page of the document manually. It shortens
                                  % the textheight of the last page by a suitable amount.
                                  % This command does not take effect until the next page
                                  % so it should come on the page before the last. Make
                                  % sure that you do not shorten the textheight too much.

\section*{Acknowledgment}

This work is supported by the National Natural Science Foundation of China (No. 62173249) and the Natural Science Foundation of Shanghai (No. 20ZR1460100).

%%%%%%%%%%%%%%%%%%%%%%%%%%%%%%%%%%%%%%%%%%%%%%%%%%%%%%%%%%%%%%%%%%%%%%%%%%%%%%%%
%%%%%%%%%%%%%%%%%%%%%%%%%%%%%%
% Reference
%%%%%%%%%%%%%%%%%%%%%%%%%%%%%%

\bibliographystyle{IEEEtran}
\normalem
\bibliography{IROS2022}

\end{document}